\documentclass[runningheads]{llncs}
\usepackage{graphicx}
\usepackage{amsmath, amssymb} 
\usepackage{ruler}
\usepackage{color}
\usepackage[width=122mm,left=12mm,paperwidth=146mm,height=193mm,top=12mm,paperheight=217mm]{geometry}
\usepackage{hyperref}
\begin{document}
\pagestyle{headings}
\mainmatter
\def\ECCV18SubNumber{}  

\title{Small-scale Pedestrian Detection Based on Somatic Topology Localization and Temporal Feature Aggregation} 

\titlerunning{Tao Song, Leiyu Sun, Di Xie, Haiming Sun, Shiliang Pu}

\authorrunning{}

\author{Tao Song, Leiyu Sun, Di Xie, Haiming Sun, Shiliang Pu}
\institute{Hikvision Research Institute}

\maketitle

\begin{abstract}
A critical issue in pedestrian detection is to detect small-scale objects that will introduce feeble contrast and motion blur in images and videos, which in our opinion should partially resort to deep-rooted annotation bias. Motivated by this, we propose a novel method integrated with somatic topological line localization (TLL) and temporal feature aggregation for detecting multi-scale pedestrians, which works particularly well with small-scale pedestrians that are relatively far from the camera. Moreover, a post-processing scheme based on Markov Random Field~(MRF) is introduced to eliminate ambiguities in occlusion cases. Applying with these methodologies comprehensively, we achieve best detection performance on Caltech benchmark and improve performance of small-scale objects significantly~(miss rate decreases from 74.53\% to 60.79\%). Beyond this, we also achieve competitive performance on CityPersons dataset and show the existence of annotation bias in KITTI dataset.
\end{abstract}

\section{Introduction}
Pedestrian detection is a critical problem in computer vision with significant impact on a number of applications, such as urban autonomous driving, surveillance and robotics. In recent years many works have been devoted to this detection task \cite{PDSummary,FeaPyramid,FRCNNPD}, however, there still leaves a critical bottleneck caused by various scales of pedestrians in an image \cite{SAFRCNN,MSCNN}. Despite current detectors work reasonably well with large-scale pedestrians near the camera, their performance always sustains a significant deterioration in the presence of small-scale pedestrians that are relatively far from the camera.

Accurately detecting small-scale pedestrian instances is quite difficult due to the following inherent challenges: Firstly, most of the small-scale instances appear with blurred boundaries and obscure appearance, thus it is hard to distinguish them from the background clutters and other overlapped instances. Secondly and more insightfully, existing methods(e.g., Faster-RCNN \cite{FRCNN}, R-FCN \cite{RFCN}) heavily rely on bounding-box based annotations, which inevitably incorporates parts of false positives(e.g., background pixels that usually occupy more than half of the rectangular area), introducing ambiguities and uncertainties to confuse classifiers. This issue is more pronounced for small-scale pedestrian instances as they retain much less information compared with large-scale instances, thus the signal to noise ratio~(SNR) is considerably decreased. In most related works \cite{FRCNNPD,SAFRCNN,MSCNN} that aim to detect small-scale objects, one will \emph{ONLY} resort to the perceptive fields of convolution. However, in our opinion, what impacts the performance of small-scale objects other than perceptive fields may reside in the very initial phase of machine learning pipeline, which is to say, the annotation phase.

On the other hand, according to the causal modeling idea proposed by \cite{CounterFact}, if one wonders whether there is a bias in bounding-box based annotations, he must figure out corresponding counterfactual: would the performance still be identical or even improved what if we had \emph{NOT} applied bounding-box based annotations?
\begin{figure}[t]
	\centering
	\includegraphics[height=4.8cm]{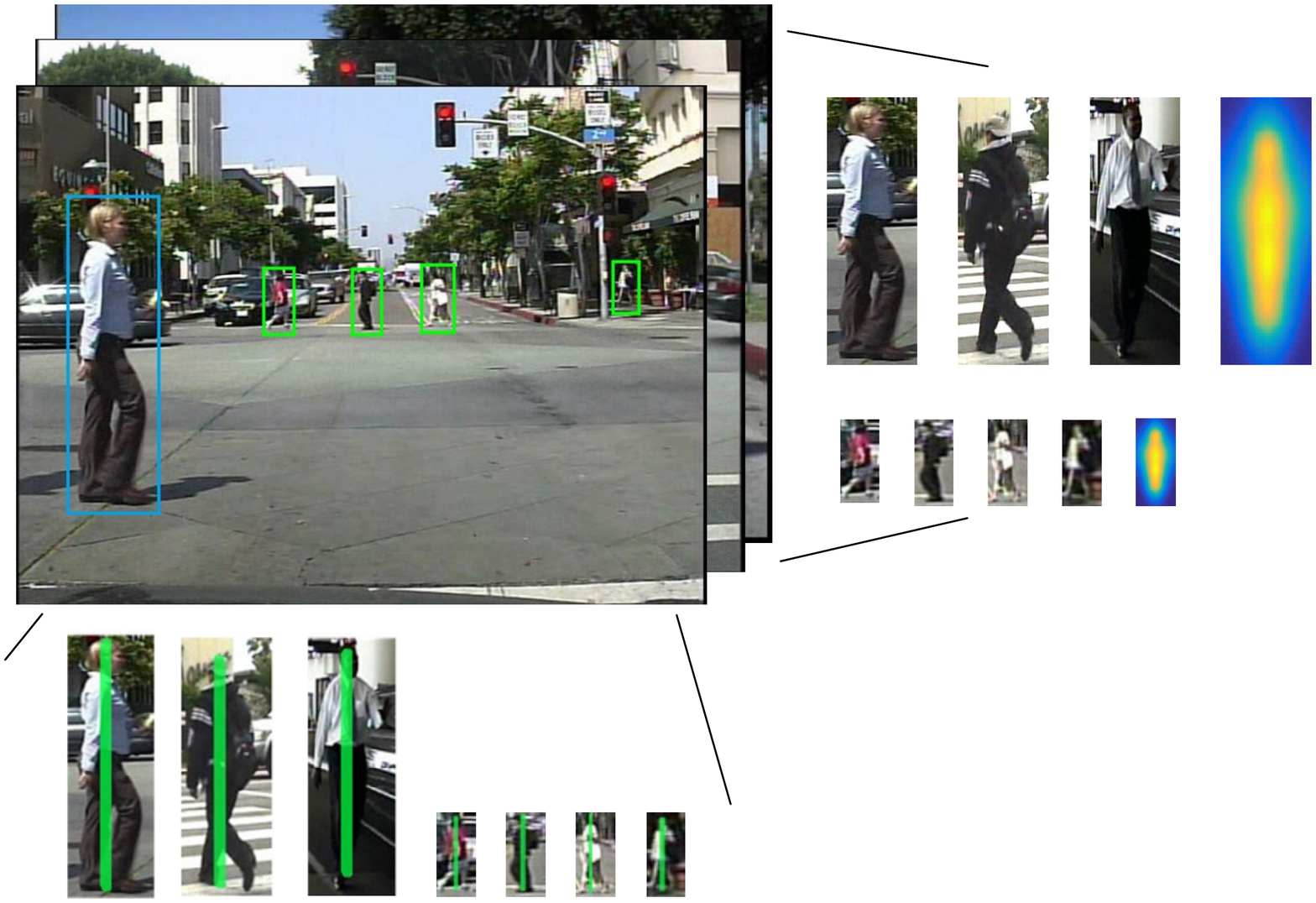}
	\caption{Pedestrians over different scales could be modeled as a group of 2D Gaussian kernels, indicating that the top-bottom topological line possess high certainty. Our approach attempt to locate this topological line for pedestrian detection.}
	\label{fig1}
\end{figure}

Motivated by above insight and counterfactual argument, we aim to address the scale variation problem with an alternative annotation, by simply locating the somatic topological line of each pedestrian as illustrated in Fig.~\ref{fig1}. This top-bottom topology is proposed due to the following consideration factors: Firstly, human bodies of various scales could be modeled as a group of 2D Gaussian kernels with different scale variances \cite{CC,SICC}. It intuitionally supposes that pixels on the top-bottom topological centre line of a human body possess high certainty, while pixels close to pedestrian contour have relatively low confidence. This hypothesis especially aligns well with the fact that small-scale instances sustain blurred boundaries and obscure appearance. Secondly, body skeletons of large-size instances, which demonstrate the detailed topology of human bodies, can provide rich information for pedestrian detection \cite{CMUPose,GooglePose,HIKPose}. However, 1)~skeletons for small-scale instances cannot be recognized easily and 2)~annotations of all the datasets are almost bounding-box, which is labor-intensive to transform them into skeletons. On the contrary, the proposed top-bottom topological line is a trade-off pivot to fuse the advantages of both automatic annotation generation and uncertainty elimination. Lastly, a simple but effective subjective test shows that compared with bounding-box based annotation, the proposed topological line demonstrates a much more consistency between annotators, especially for the small-scale instances as shown in Sec.~\ref{sec:anno_com}.

On basis of the topological line annotation, we devise a fully convolutional network (FCN) that takes multi-scale feature representations and regresses the confidence of topological elements, i.e., top and bottom vertex, as well as the link edge between them. To eliminate ambiguous matching problem in crowded cases, a post-processing scheme based on Markov Random Field (MRF) is introduced to keep each predicted instance away from the other predicted instance with different designated objects, making the detection results less sensitive to occlusion. Moreover, we design a scheme to utilize temporal information by aggregating features of adjacent frames to further improve performance. Empirical evaluation reveals the novel TLL networks with or without temporal feature aggregation both lead to state-of-the-art performance on Caltech~\cite{Caltech} and CityPersons~\cite{CityPersons} datasets.

In summary our key contributions are as follows:
\begin{itemize}
	\item From the counterfactual view, we attempt to prove that topological annotation methodologies other than bounding box will introduce less ambiguity, which results in better performance and is especially effective for small-scale objects. Meanwhile, the deep-rooted bounding-box based annotation bias is challenged by our work, which is thought-provoking to rethink how to provide classifiers with discriminative information.
	\item We devise a unified FCN based network to locate the topological somatic line for detecting multi-scale pedestrian instances while introduce a post-processing scheme based on MRF to eliminate ambiguities in occlusion cases. A temporal feature aggregation scheme is integrated to propagate temporal cues across frames and further improves the detection performance.
	\item To the best of our knowledge, we achieve best detection performance on Caltech benchmark and improve performance of small-scale objects significantly~(miss rate decreases from 74.53\% to 60.79\%). On CityPersons dataset, our proposed method obtains superior performance in occlusion cases without any bells and whistles. Beyond these, the existence of annotation bias in KITTI dataset is disclosed and analyzed.
\end{itemize}

\section{Related Work}
\subsection{Multi-scale Object Detection}
State-of-the-art methods for multi-scale object detection are mainly based on the pipeline of classifying region proposals and regressing the coordinates of bounding boxes, e.g., Faster-RCNN \cite{RCNN,FRCNN,RFCN}, YOLO \cite{YOLO1,YOLO2} and SSD \cite{SSD}. RPN+BF method \cite{FRCNNPD} uses boosted forests classiﬁers on top of the region proposal network (RPN) and high-resolution convolutional features to effective bootstrapping for mining hard negatives. SA-FastRCNN \cite{SAFRCNN} develops a divide-and-conquer strategy based on Fast-RCNN that uses multiple built-in subnetworks to adaptively detect pedestrians across scales. Similarly, \cite{MSCNN} proposes a uniﬁed multi-scale convolutional neural network (MS-CNN), which performs detection at multiple intermediate layers to match objects of different scales, as well as an upsampling operation to prevent insufficient resolution of feature maps for handling small instances. Rather than using a single downstream classifier, the fused deep neural network (F-DNN+SS) method \cite{FDNN} uses a derivation of the Faster R-CNN framework fusing multiple parallel classifiers including Resnet \cite{ResNet} and Googlenet \cite{GoogleNet} using soft-rejection, and further incorporates pixel-wise semantic segmentation in a post-processing manner to suppress background proposals. Simultaneous Detection \& Segmentation RCNN (SDS-RCNN) \cite{SDSRCNN} improves object detection by using semantic segmentation as a strong cue, infusing the segmentation masks on top of shared feature maps as a reinforcement to the pedestrian detector. Recently, an active detection model (ADM) \cite{ADM} based on multi-layer feature representations, executes sequences of coordinate transformation actions on a set of initial bounding-box proposals to deliver accurate prediction of pedestrian locations, and achieve a more balanced detection performance for different scale pedestrian instances on the Caltech benchmark. However, the aboved bounding-box based methods inevitably incorporates a large proportion of uncertain background pixels (false positive) to the human pattern, while impels the instances to be predicted as false negatives. In practice, it may lead to compromised results with particularly poor detections for small-scale instances. On the contrary, our approach relies on locating the somatic topology with high certainty, which is naturally flexible to object scale and aspect ratio variation.
\subsection{Line Annotation}
Line annotation is first proposed in \cite{NewCal,CityPersons} to produce high-quality bounding-box ground truths(GTs). The annotation procedure ensures the boxes align well with the center of the subjects, and these works show that better annotations on localisation accuracy lead to a stronger model than obtained when using original annotations. However, best results of these work are achieved on the validation/test set with a sanitised version of annotations, which is unfair when compared with other advanced methods evaluated on the original annotation set. What's more, the work in \cite{NewCal} shows that models trained on original/new and tested on original/new perform better than training and testing on different annotations. In contrast, our work utilizes the line annotation in a different way: the line annotation is not used to produce bounding-box GTs, but GTs themselves, and we design a FCN to regress the topological elements of the line. Meanwhile, tight bounding-boxes with a uniform aspect ratio could be automatically generated from each predicted topological lines and the detection results could be evaluated on the original annotation, which leads a fair comparison with the state-of-the-art methods.
\subsection{Temporal Feature Aggregation}
Temporal cues could be incorporated for feature reinforcement in object detection tasks. For example, TCNN \cite{TCNN} uses optical flow to map detections to neighboring frames and suppresses low-confidence predictions while incorporating tracking algorithms. FGFA \cite{FGFA} improves detection accuracy by warping and averaging features from nearby frames with adaptive weighting. However, its flow-subnet is trained on synthetic dataset \cite{FlowNet}, which obstructs itself from obtaining optical flow accurately in real scenes. A recent work, \cite{VOD} creates a recurrent-convolutional detection architecture by combining SSD \cite{SSD} with LSTM, and designs a bottleneck structure to reduce computational cost. Inspired by the above ideas, we unify the proposed TLL with recurrent network into a single temporally-aware architecture.
\section{Annotation Comparison}\label{sec:anno_com}
\begin{figure}[t]
	\centering
	\includegraphics[height=3.6cm]{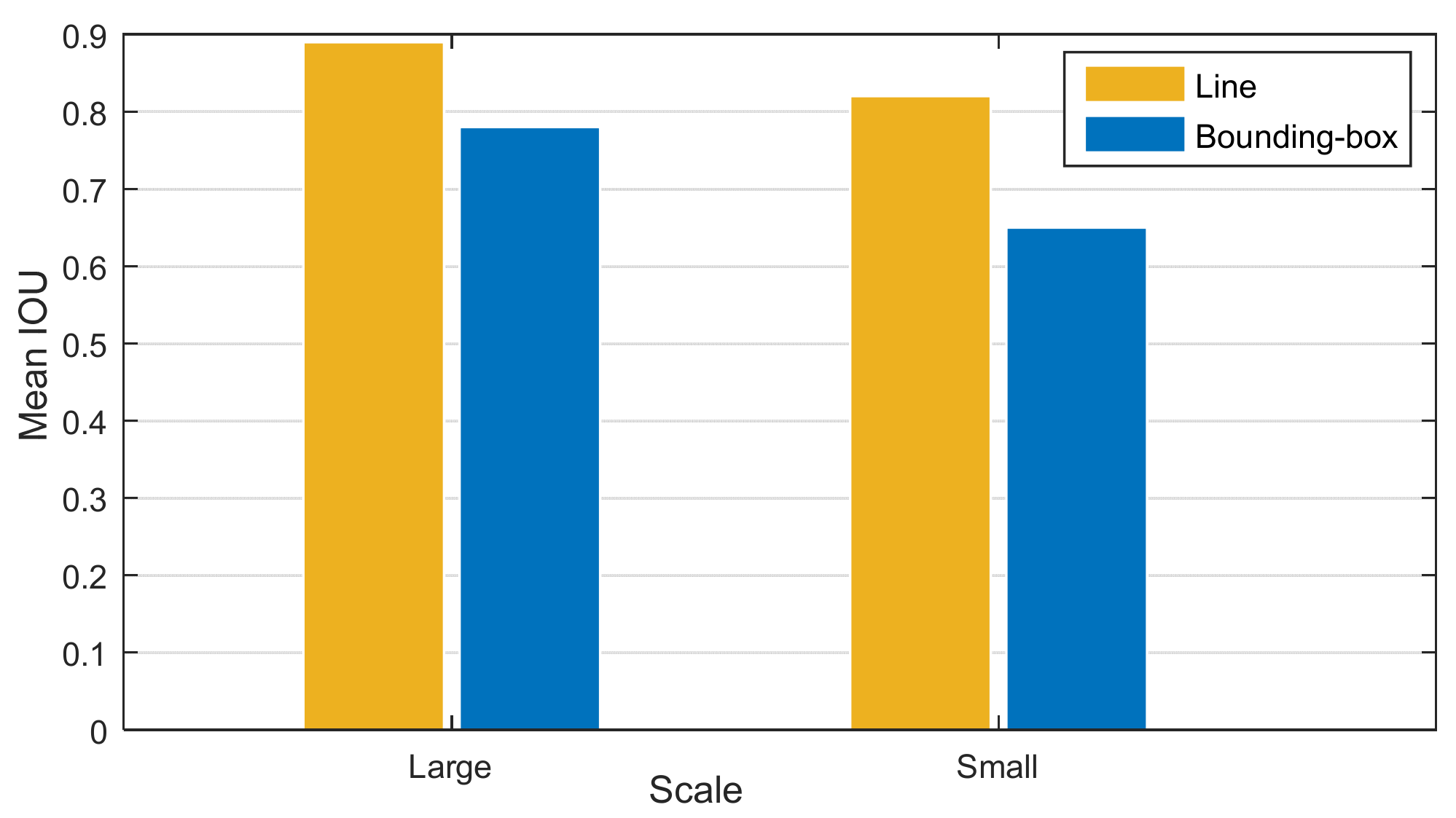}
	\caption{Mean-IoUs comparison of two annotation for different scale pedenstrians.}
	\label{fig2}
\end{figure}

To compare the line and bounding-box annotation methods, we design a simple subjective test. We extract 200 independent frames containing multi-scale pedestrians from Caltech training video-data, and hire 10 annotators to produced duplicate annotations via the two annotation methods separately. In each round, each annotator is shown the set of frames in random order and draws pedestrian instances by one annotation measure with a label tool. Annotators are asked to hallucinate head and feet if they are not visible. After that, pedestrian instances annotated by all 10 annotators are collected for evaluation. This procedure is indispensable since it's unreasonable to request each annotator exhaustedly outlines all, and the same instances from each image under the situation that many small-scales, defocus or blurred instances exist. Then we assess the two annotations using IoU (intersection over union) calculated between the overlap of 10 annotations and the union of them. Following \cite{NewCal}, bounding-boxes with uniform aspect ratio could be automatically generated such that its centre coincides with the centre point of the manually-drawn axis. In Fig.~\ref{fig2}, we compare the mean IoUs of two annotations for large-scale (pedestrian height $\geq$ 80 pixels) and small-scale (pedestrian height $<$ 80 pixels) pedestrians. Note the bounding-box annotation instances are normalized to the same aspect ratio as line annotation ones for fair statistics.

The test result emphasizes that line annotation promotes more precise localisation on pedestrian than marking a bounding box, especially for small-scale instances. The reason lies in that annotators tend to align well with the center of subjects when drawing lines. While for the small-scale cases, even a few pixels mismatch on the bounding box annotation results in low IoUs, thus line annotation has a much lower variation compared with bounding-box. Besides, this test also tells us all the annotation methodologies are subjective and bounding-box based ones are prone to produce bias as shown in Fig.~\ref{fig7}(a), which confuses any classifiers to deteriorate performance.
\section{TLL Detection Methodology}
\label{sec:proposed}

In this section, we describe the TLL detector for multi-scale pedestrians. As the core of our work, we firstly describe the single-shot network that regresses somatic topological elements. Then we discuss how to utilize the multi-scale representational features within the network, and employ the MRF scheme for dealing with crowd occlusion. Finally, the scheme of integrating TLL with temporal information for further detection improvement will be introduced.

\subsection{Topological Line Localization Network}
\begin{figure}[t]
	\centering
	\includegraphics[height=4.8cm]{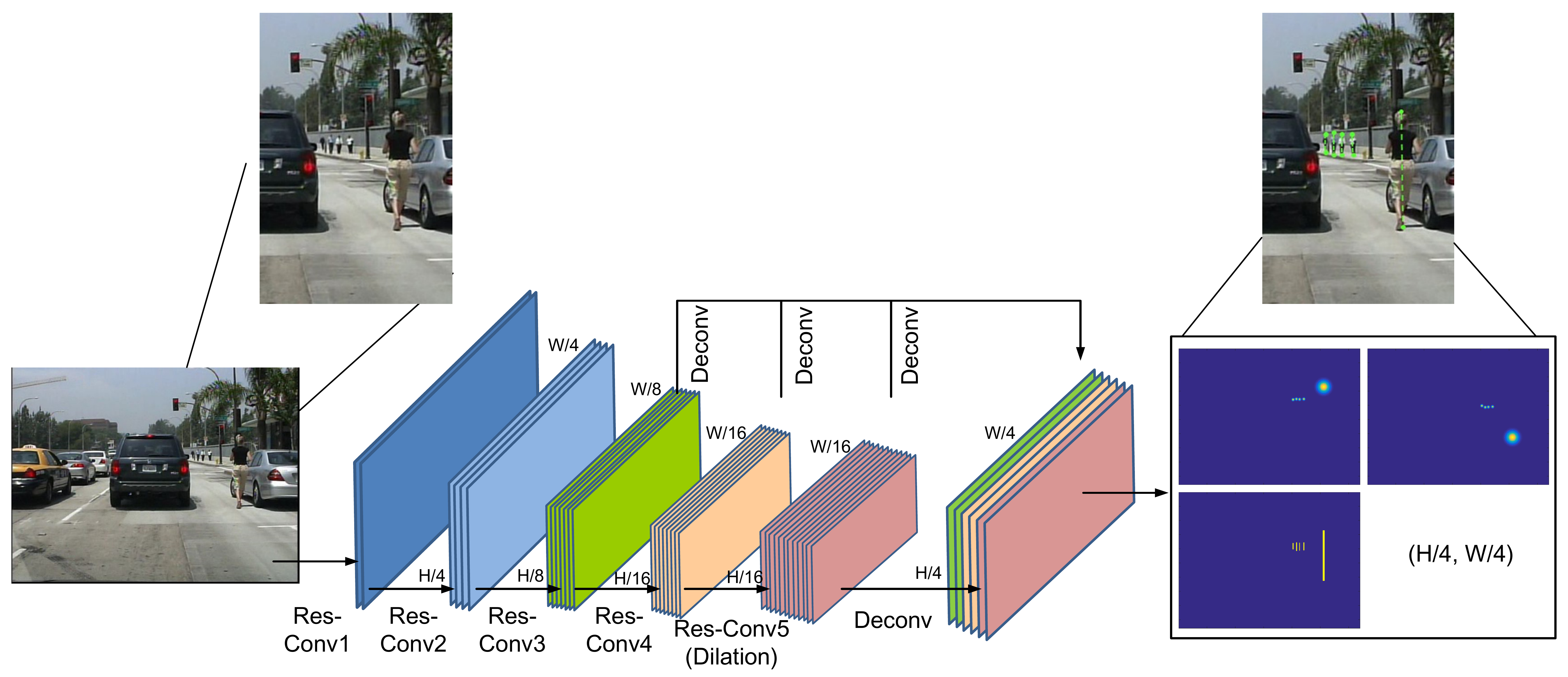}
	\caption{An overview of single-shot TLL detection network.}
	\label{fig3}
\end{figure}

An overview of the single-shot TLL detection scheme is depicted in Fig.~\ref{fig3}. The backbone of TLL is a Resnet-50 network, which is fast and accurate for object recognition \cite{ResNet}. We extend it to a fully convolutional version for an input image of arbitrary size, by using series of dilated-convolution, deconvolution, and skip connection methods. Specifically, as the default network has a feature stride of 32 pixels, which is too large to localize small-scale pedestrians, thus we remove the down-sampling in Conv5x and use dilated-convolution for keeping the receptive field, resulted in the final feature map as 1/16 of input size. Following the representation theory, higher layer features tend to encode more global and semantic information of objects that is robust against appearance variations, while outputs of lower layers provide more precise localization. We extract features from the last layer of each res-block started from Conv3 (i.e., Resnet50-Conv3d, Conv4f, Conv5c, detailed in Sec.~\ref{sec:multi-rep}.) and recover their spatial resolutions to 1/4 of the input size using deconvolution. These multi-layer representations are skip connected for regressing the top and bottom vertex confidence maps, as well as the map of link edge between them.

Every top and bottom vertex locations are modeled as a Gaussian peak. Let $p_{k}$ be the ground-truth (GT) top/bottom vertex positions of {\it i}-th pedestrian in the image, then the GT vertex confidence map $D(x)$, is formed by max aggregation of all ${N_{k}}$ pedestrian peaks in the image.

\begin{align}
	D(x)=\max\limits_{k \in {N_{k}}} d(x; p_{k},\sigma)
\end{align}
where $x \in R^{2}$ stands for one pixel location in the confidence map, and $d(x; p,\sigma)$ is a two dimension Gaussian distribution with empirically chosen variance $\sigma$.

Link edge of a pedestrian $l(\textbf{x})$ is modeled as a connecting line between the two vertexes, with a width scaled by the height of pedestrian. Pixel values of the line are given as a unit vector {\bf v} in the direction from the top vertex pointing to the bottom vertex. Thus the GT link value map $L(x)$ is an average of all ${N_{k}}$ pedestrian links in the image.
\begin{align}
	L(x) &= \frac{1}{N_{k}} \sum\limits_{k \in {N_{k}}} l_{k}(x)
\end{align}
where ${N_{j}}$ is the total number of pedestrians within an image, and $l(x)$ is defined as:
\begin{align}
	l(x) &= \left\{
	\begin{array}{lr}
		{\bf v} \text{  if \textit{x} on link edge of a pedestrian} & \\
		0 \text{  otherwise} &
	\end{array}
	\right.
\end{align}

During training,  mean squared error (MSE) is used to measure the difference between the predicted conﬁdence maps and GT. The loss function $f$ is deﬁned as follows:
\begin{align}
	f &=\Arrowvert\tilde{D_{t}}-D_{t}\Arrowvert_{2}^{2} + \Arrowvert\tilde{D_{b}}-D_{b}\Arrowvert_{2}^{2} + \lambda \Arrowvert\tilde{L}-L\Arrowvert_{2}^{2}
\end{align}
where  $\tilde{D_{t}}$ and $\tilde{D_{b}}$ stand for the predicted vertex conﬁdence maps, $\tilde{D_{b}}$ stands for the predicted link map, and $\lambda$ is a weighting factor that balances the vertex confidence error and link confidence error. 

During inference, given an image {\it I}, candidate top and bottom vertex locations, $\tilde{t}_{i}$ and $\tilde{b}_{j}$, could be located by performing non-maximum suppression (NMS) on the predicted vertex conﬁdence maps $\tilde{D_{t}}$ and $\tilde{D_{b}}$. Then link score of each edge candidate (i.e., each pair of possible connections between candidate top and bottom vertexes) is calculated by measuring the alignment of the predicted link $\tilde{L}$ with the candidate edge that formed by connecting the candidate top and bottom vertexes.
\begin{align}
	E_{i, j} &= \int_{u=0}^{u=1} \tilde{L}(p(u))\cdot \dfrac{\tilde{b}_{j}-\tilde{t}_{i}} {\Arrowvert\tilde{b}_{j}-\tilde{t}_{i}\Arrowvert_{2}}du
\end{align}
where ${\it p(u)}$ indicates the sampling points along the edge candidate. Then based on the maximum confidence scores of each edge candidates, finding the top-bottom vertex pairs becomes a bipartite graph matching (BGM) \cite{BGM} problem which could be easily solved by the Hungary algorithm \cite{Hungarian}. Thus the predicted link of one vertex pair is determined as the topological line location of a pedestrian, with a detection score calculated by multiplication of vertex and link confidences.

\subsection{Multi-scale Representation}\label{sec:multi-rep}

\begin{figure}[t]
	\centering
	\includegraphics[height=5.5cm]{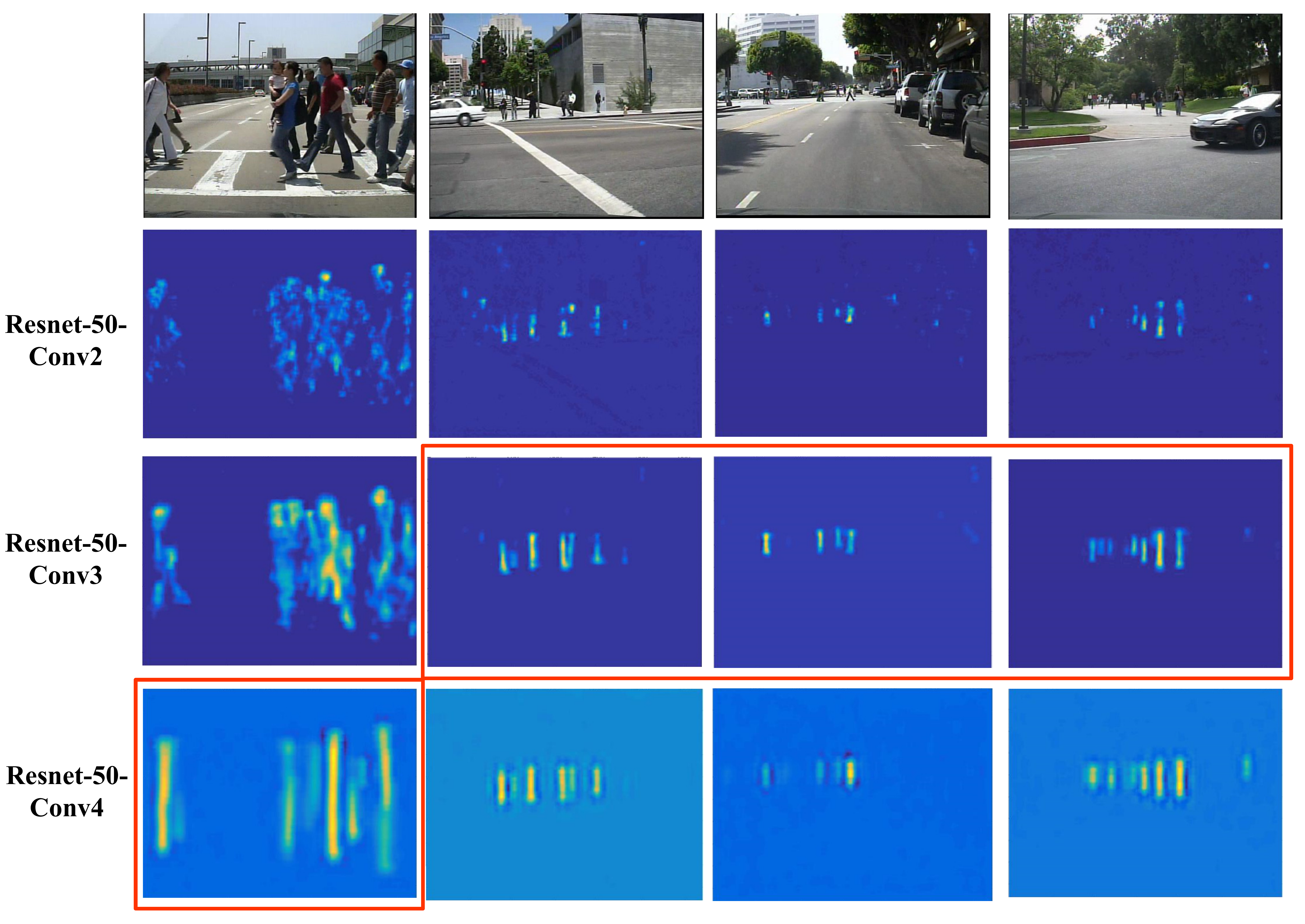}
	\caption{Visualization of the predicted link maps from different intermediate layers. Red bounding boxes indicate the optimal activations across multiple representation.}
	\label{fig4}
\end{figure}

It has been revealed in \cite{SAFRCNN,ADM} that large-scale pedestrian instances typically exhibit dramatically different visual characteristics and internal features from the small-scale ones. For the network, pedestrian instances of different scales should have different responses at distinct feature representation layers. To investigate the optimal start of res-block features for skip connection in our network, we regress the confidence maps directly from different intermediate feature maps to visualize the responses at different convolutional layers for detecting pedestrians of various scales. Fig.~\ref{fig4} illustrates the regressed link confidence maps from three intermediate convolutional layers, i.e., Resnet50-Conv2, Conv3 and Conv4.

In general, convolutional features are effective only at a proper scale where optimal activation is obtained. Lower representation layers have a strong response for small-scale pedestrians, while large-scale pedestrian instances are usually detected by higher layers. Specifically, small-scale pedestrians are most comfortably picked up at Conv3, and large-scale ones are largely detected at Conv4. Interestingly, the much lower layer Conv2, does not have strong responses for each scale instances, the reason may due to its primitive and limited semantic characteristics. In practice, we choose Conv3 as a satisfactory starting layer for effective multi-scale representation.

\subsection{MRF based Matching}

As presented above, BGM produces detection results depending on the maximum link scores of candidate top and bottom vertex pairs. Whereas, in crowd scenes, pedestrians often gather together and occlude each other. This may cause the TLL network to output very close vertices, and high link scores between each candidate pairs, leading to confused detection results. Thus, the crowd occlusion severely increases the difficulty in pedestrian localization. In order to robustly localize each individual pedestrian in crowd scenes, locations of its surrounding objects need to be taken into account.
\begin{figure}[t]
	\centering
	\includegraphics[height=3.1cm]{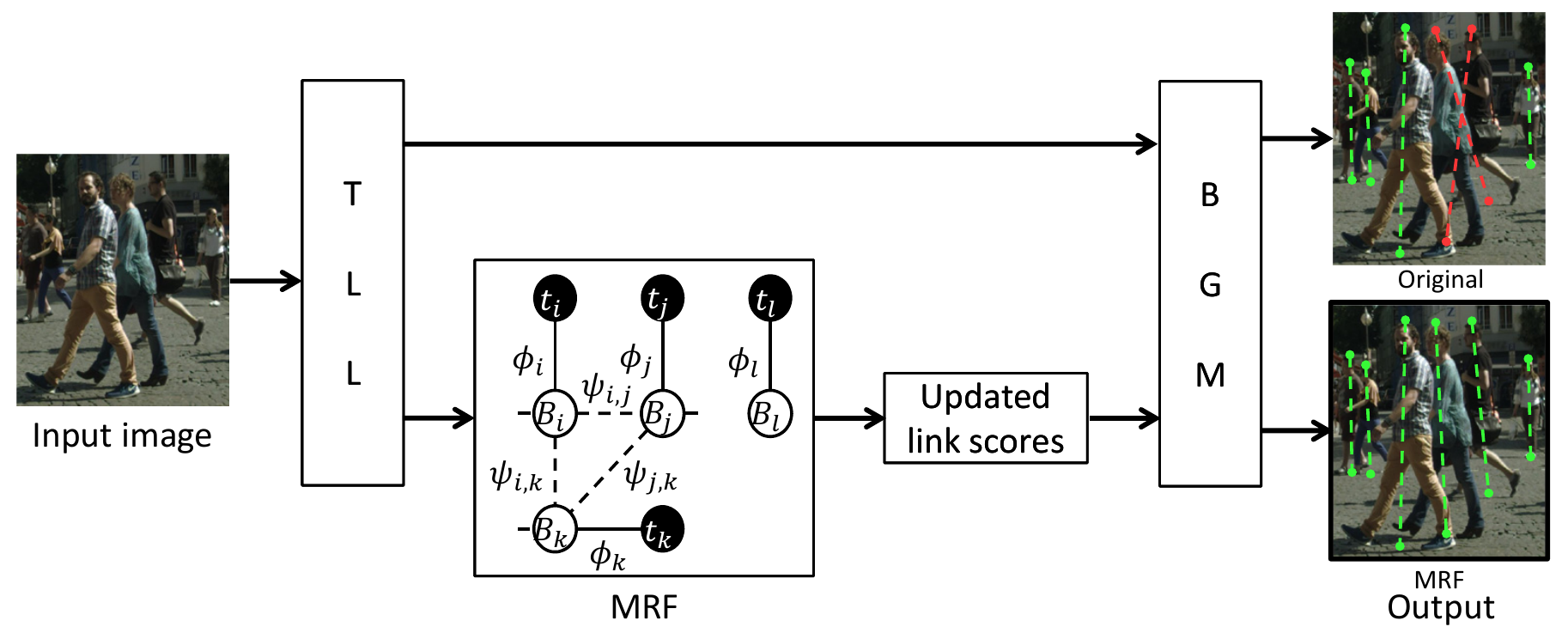}
	\caption{MRF based matching compared with the original method. Note the dotted lines in red represents the mismatches under occlusion cases.}
	\label{fig5}
\end{figure}

A MRF based post-processing scheme is constructed as shown in Fig.~\ref{fig5}. For one candidate top vertex $t_{i}$, there exists several (e.g., $N_i$) candidate bottom vertices whose link scores are high and close due to occlusion, denoted as ${B_i} = \left\{ {b_n^i} \right\}_{n = 1}^{{N_i}}$. This candidate top vertex and its corresponding bottom vertices form a subset, and they are designated as the observed node and hidden node respectively. Link scores ${E_i} = \left\{ {e_n^i} \right\}_{n = 1}^{{N_i}}$ between $t_{i}$ and ${B_i}$ are set as the joint compatibility ${\phi _i}\left( {{t_i},{B_i}} \right)$. For each candidate top-bottom vertex pair $\left\{ {{t_i},b_n^i} \right\}$ within a subset, one virtual bounding-box is automatically generated with a uniform aspect ratio, forming $VB_{i}$. Then IoUs of every two virtual boxes from two different subsets could be calculated. The IoUs between two subsets reflect a neighboring relationship between them. The extent of two neighboring subset \textit{i} and \textit{j} away from each other is set as the neighboring compatibility: 
\begin{align}
{\psi _{i,j}({B_i},{B_j})} = {\rm{exp}}\left( { - IoU\left( {V{B_i},V{B_j}} \right)/\alpha } \right)
\end{align}
where $\alpha$ is a normalization parameter. Max-product algorithm is utilized to optimize the objective function:
\begin{align}
\mathop {\min }\limits_B {\rm{p}}\left( {\left\{ B \right\}} \right) = \frac{1}{Z}\mathop \prod \limits_{\left( {i,j} \right)} {\psi _{i,j}}\left( {{B_i},{B_j}} \right)\mathop \prod \limits_i {\phi _i}\left( {{t_i},{B_i}} \right)
\end{align}
where $Z$ is a normalization constant. After several iterations, MRF converges and optimized confidences ${C_i} = \left\{ {c_n^i} \right\}_{n = 1}^{{N_i}}$ of hidden node ${B_i}$ could be obtained. Then link scores of the candidate vertex pairs $\left\{ {{t_i},b_n^i} \right\}$ are updated: 
\begin{align}
s_n^i = c_n^i*\mathop \sum \nolimits_n^{{N_i}} e_n^i
\end{align}
Finally, BGM is utilized to generate detection results on the basis of updated link scores. The MRF adds an extra constraint that pushes top-bottom vertex pairs away from each other, leading to less mismatches under occlusion cases.

\subsection{Multi-frame Temporal Feature Aggregation}
\begin{figure}[t]
	\centering
	\includegraphics[height=4.4cm]{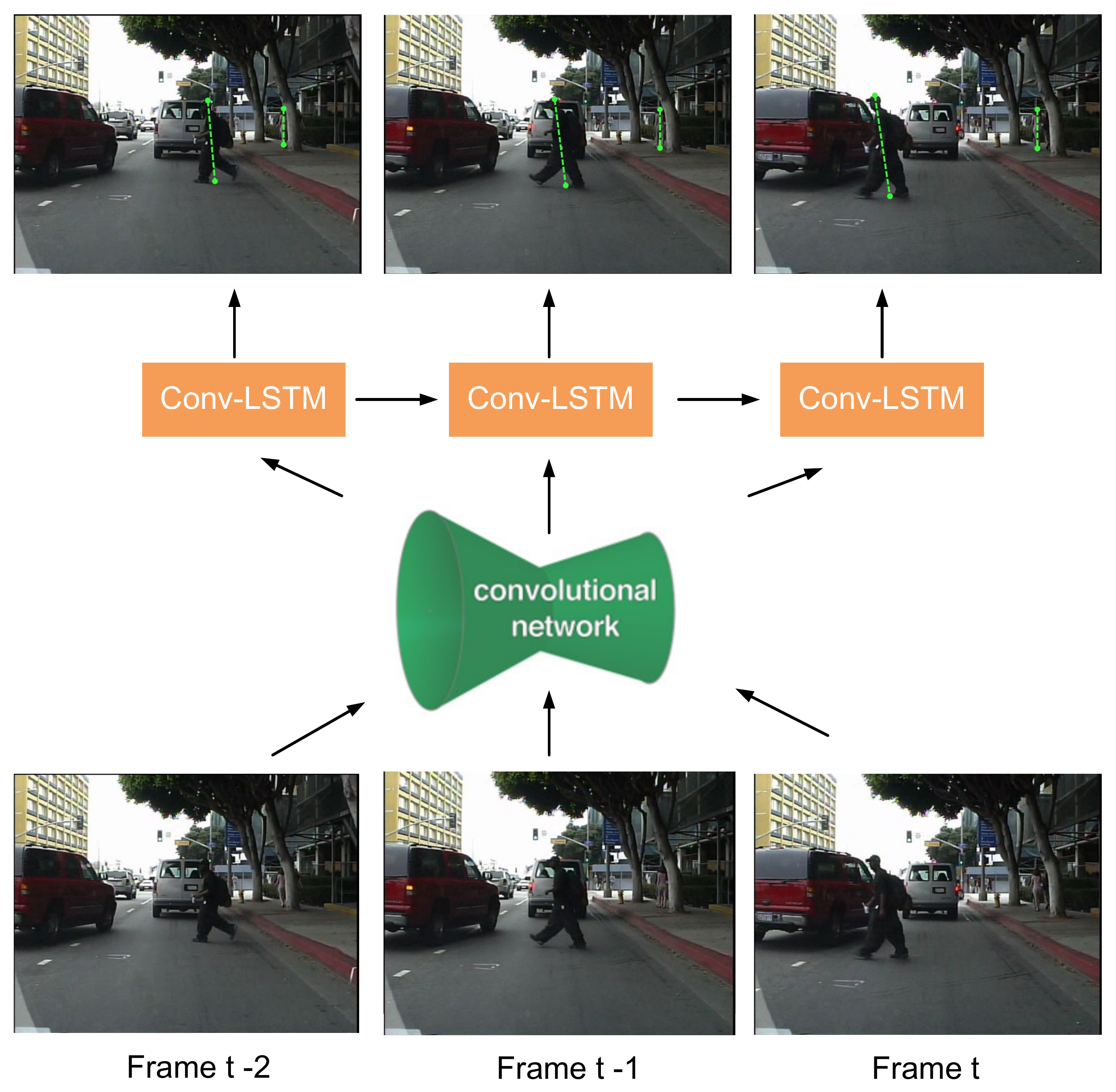}
	\caption{Illustration of our joint TLL+Conv-LSTM model.}
	\label{fig6}
\end{figure}
We seek to improve the detection quality by exploiting temporal information when videos are available. RNN has been verified as an effective way to make use of motion information \cite{VC,VAR}. Thus we try to unify the proposed TLL and recurrent network into a single temporally-aware architecture. Conv-LSTM \cite{ConvLSTM} is incorporated as a means of propagating frame-level information across time. Specifically, for a video sequence, convolutional layers for representation are shared by each frame to extract spatial features, then multi-layer features of each frame are taken as input to the Conv-LSTM layer. At each time step, it refines output features on the basis of the state and input, extracts additional temporal cues from the input, and updates the state. Then outputs of Conv-LSTM are utilized for further regression. An illustration of our joint TLL+Conv-LSTM model can be seen in Fig.~\ref{fig6}. Comparing with FGFA~\cite{FGFA}, Conv-LSTM implicitly aggregates feature maps in a more comprehensive way, which overcomes the feature scatter disadvantage of pixel-wise aggregation.

\section{Experiments}
\subsection{Experiment Settings}
\begin{figure}[t]
	\centering
	\includegraphics[height=3.6cm]{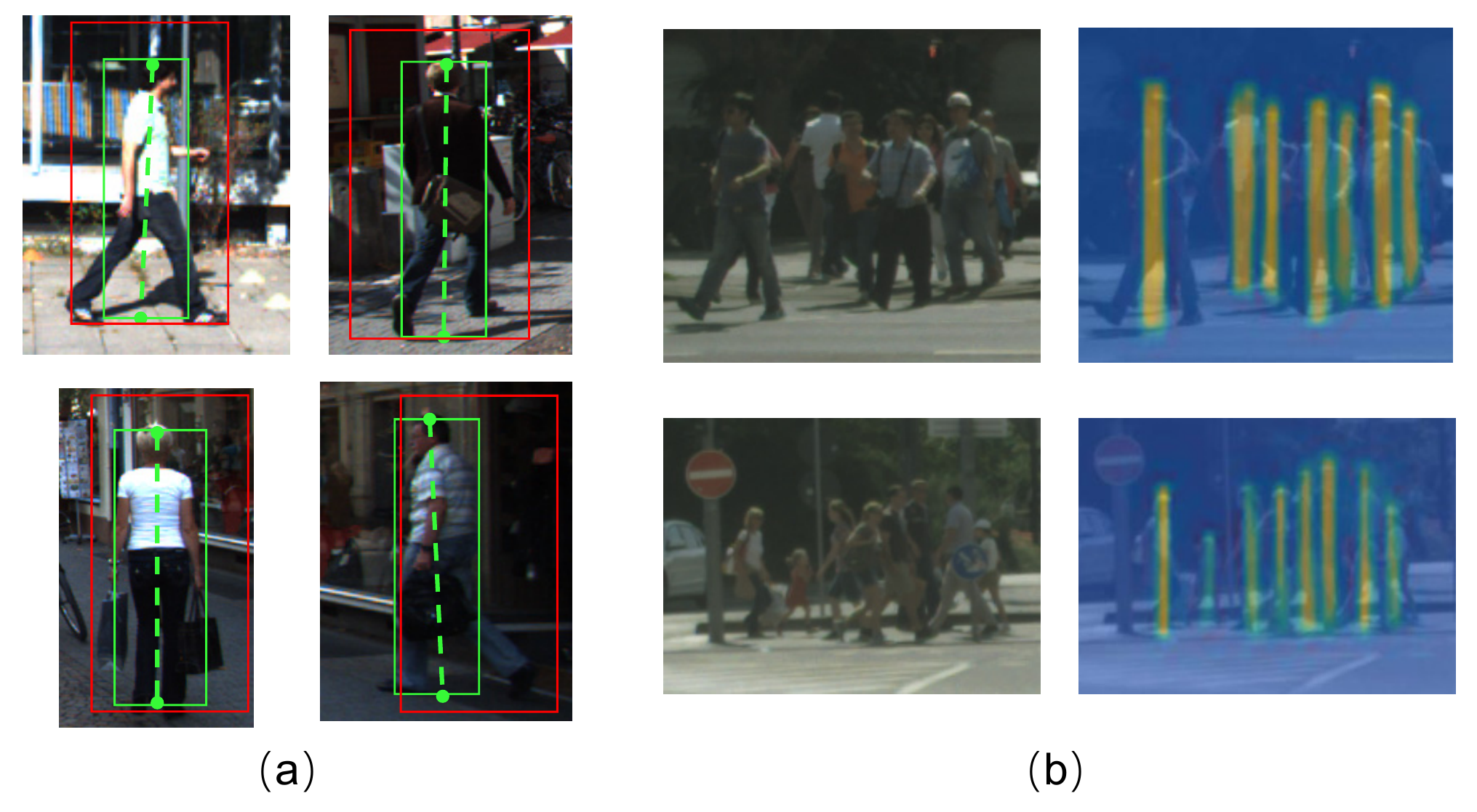}
	\caption{(a) Samples of our detection results (green) and GT boxes (red) in KITTI validation set. (b) The effectiveness of our predicted link map in crowd scenes.}
	\label{fig7}
\end{figure}

We examine the proposed approach on widely used benchmarks including Caltech \cite{Caltech} and Citypersons \cite{CityPersons}. We follow the standard evaluation metric: log miss rate is averaged over the false positive per image (FPPI) in [$10^{{-}2}, 10^{0}$], denoted as MR. These datasets provide different evaluation protocols on the basis of annotation sizes and occlusion levels. One reason for choosing the two datasets is that they provide tight annotation boxes with normalized aspect ratio, thus the TLL detection results could be fairly evaluated on the original annotation, and compared with other state-of-the-art methods. Specifically, the top-down centre axis of each annotated bounding-box is used as the approximate GT topological line for training, which leads to no burden of additional human annotation on these datasets. From each predicted topological line, a tight bounding-box with uniform aspect ratio (0.41) could be automatically generated such that its centroid coincides with the centre point of the topological line. Then IoUs between this bounding-box and GT boxes could be calculated for the evaluation process. Tight and normalized annotation is important for the quantitative results during evaluation, as one correct detection may suffer a low IoU with its GT box with irregular aspect ratio (such as walking persons have varying width, as shown in Fig.~\ref{fig7}(a)), resulted in a false positive to pull down \textit{Precision}, while a false negative to pull down \textit{Recall}, which results in only $38.72 \%$ average precision for the moderate test set on KITTI dataset \cite{KITTI} and in a sense reveals the annotation bias introduced by subjective judgement.

Every frame in Caltech has been densely annotated with the bounding boxes of pedestrian instances. This dataset is unique to others for following reasons: First, over 70\% of the annotated pedestrian instances have a height smaller than 100 pixels, including extremely tiny instances under 50 pixels, which is rare for other datasets. Second, the dataset provides original videos, on which our multi-frame aggregation methods could be evaluated. The standard test set of 4024 images is used for evaluation under different protocols.

CityPersons is a new pedestrian detection dataset on top of the semantic segmentation dataset CityScapes \cite{CityScapes} and consists more crowded scenes compared with Caltech, and over 20\% of pedestrian annotations overlap with another annotated pedestrian whose IoU is above 0.3. As CityPersons provides image samples only, its validation set with 500 images is used for the evaluation of our single-shot network.

\subsection{Single-shot Network}
For the Caltech benchmark, following the typical protocol in literatures \cite{FRCNNPD,SAFRCNN,MSCNN}, we use dense sampling of the training data (every 3th frame, resulted in 42782 images). We compare a set of recently proposed state-of-the-art methods and quantitative results are listed in Table~\ref{Cal_res_tab}.
\setlength{\tabcolsep}{4pt}
\begin{table}[t]
	\begin{center}
		\caption{Comparison results of TLL with recent state-of-the-art methods on standard test set of Caltech (lower is better).}
		\label{Cal_res_tab}
		\begin{tabular}{lccccc}
			\hline\noalign{\smallskip}
			Methods/MR(\%) & Reasonable & All & Near & Middle & Far\\
			\noalign{\smallskip}
			\hline
			\noalign{\smallskip}
			RPN+BF \cite{FRCNNPD}  & 9.58 & 64.66 & 2.26 & 53.93 & 100\\
			SA-FastRCNN \cite{SAFRCNN}  & 9.68 & 62.59 & 0.00 & 51.83 & 100\\
			MS-CNN \cite{MSCNN}  & 9.95 & 60.65 & 2.60 & 49.13 & 97.23\\
			F-DNN+SS \cite{FDNN}  & 8.18 & 50.29 & 2.82 & 33.15 & 77.37\\
			UDN+SS \cite{UDN}  & 11.52 & 64.81 & 2.08 & 53.75 & 100\\
			SDS-RCNN \cite{SDSRCNN}  & 7.36 & 61.50 & 2.15 & 50.88 & 100\\
			ADM \cite{ADM}  & 8.64 & 42.27 & 0.41 & 30.82 & 74.53\\
			TLL  & 8.45 & 39.99 & 0.67 & 26.25 & 68.03\\
			TLL(MRF)  & 8.01 & 39.12 & 0.67 & 25.55 & 67.69\\
			\noalign{\smallskip}
			\hline
			\noalign{\smallskip}
			TLL(MRF)+FGFA \cite{FGFA}  & 7.92 & 38.58 & 0.99 & 24.39 & 63.28\\
			TLL(MRF)+LSTM  & 7.40 & 37.62 & 0.72 & 22.92 & 60.79\\	
			\noalign{\smallskip}		
			\hline
		\end{tabular}
	\end{center}
\end{table}
\setlength{\tabcolsep}{1.4pt}

\setlength{\tabcolsep}{4pt}
\begin{table}[t]
	\begin{center}
		\caption{Comparison results of single-shot TLL with recent state-of-the-art methods on validation set of Citypersons (lower is better).}
		\label{City_res_tab}
		\begin{tabular}{lcccc}
			\hline\noalign{\smallskip}
			Methods/MR(\%) & Reasonable & Heavy & Partial & Bare\\
			\noalign{\smallskip}
			\hline
			\noalign{\smallskip}
			Citypersons \cite{CityPersons}  & 15.4 & -- & -- & --\\
			Repulsion Loss\cite{RLoss}  & 13.2 & 56.9 & 16.8 & 7.6\\ 
			TLL    & 15.5 & 53.6 & 17.2 & 10.0\\
			TLL(MRF)  & 14.4 & 52.0 & 15.9 & 9.2\\
			\noalign{\smallskip}		
			\hline
		\end{tabular}
	\end{center}
\end{table}

\setlength{\tabcolsep}{1.4pt}
The proposed TLL demonstrates constantly competitive results with the state-of-the-arts, and achieves leading performance for small-scale objects (i.e., the Far and Middle protocols). Among them, performance on far-distance instances is improved most significantly, achieving a MR of 68.03\%/67.69\% with/ without MRF, which clearly exceeds the best existing results, 74.53\% of ADM, to the best of our knowledge. Middle-distance instances get an obvious gain from 30.82\% to 26.25\%/25.55\% with/without MRF as well. For the near-distance ones, as the line annotation includes more background pixels on large-scale instances, TLL does not perform better than others but a similar MR close to 0\% is achieved. Our approach outperforms the rest methods under the ALL protocol, in which significant occlusion exists. This is reasonable as the predicted link aligns well with the centre location of human body, as shown in Fig.~\ref{fig7}(b), which naturally avoids the adverse factors faced by bounding-box based methods, since crowd occlusion makes these detectors sensitive to NMS. Moreover, the MRF adjusts the matching scores in an appropriate way, resulting in an even better performance. Our proposed method can achieve improved performance based on the \emph{old} annotations, in an unprejudiced sense, without any human intervention, while \cite{RLoss} just reports performance evaluated on the \emph{new} annotations~\cite{NewCal}, which is obviously unfair to compare it with other listed methods.

For the Citypersons, we take all 3000 images in train set for training, and use the annotated bounding-boxes in a similar way as Caltech. Quantitative results are listed in Table~\ref{City_res_tab}. Since the dataset consists more crowded scenes, it can be seen that MRF acts as a more important role, and the best result is achieved in the heavy occlusion case. Interestingly, TLL alone surpasses \cite{RLoss} when people are occluded from each other heavily, which is the case \cite{RLoss} proposed to solve with. Results in Table~\ref{City_res_tab} demonstrate that it is better to provide less ambiguous information to classifiers instead of improving classifiers themselves.
\subsection{Multi-frame Feature Aggregation}
\begin{figure}[t]
	\centering
	\includegraphics[height=4.05cm]{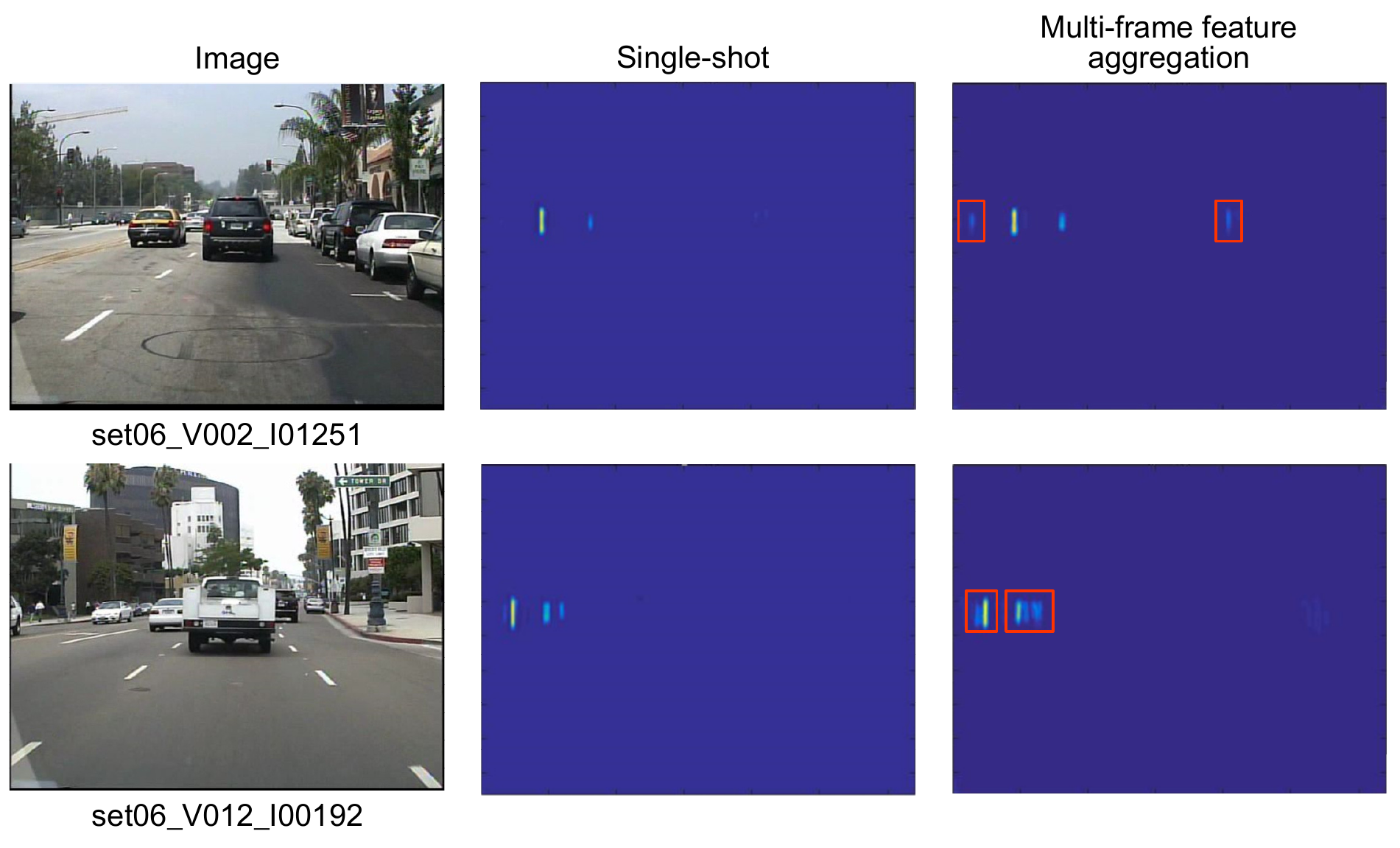}
	\caption{Visualization examples of the multi-frame feature aggregation effect. Red bounding boxes indicate the enhanced high-activated feature locations.}
	\label{fig8}
\end{figure}

Large feature channels in FCN will greatly increase the feature dimensions, bringing issues of computational overhead and memory consumption. To address this issue, we convert the multi-layer features before Conv-LSTM, down to 256 channels by using a 1$\times$1 convolutional transform. This operation is similar as the bottleneck structure in \cite{VOD}, but more efficient. After that, Conv-LSTM layers with 256 channels are inserted into the single-shot TLL network. We try to incrementally stack Conv-LSTM layers to the network, however, due to difficulties in training multiple RNNs, experiments show that stacking two or more Conv-LSTMs is not beneficial. We unroll the Conv-LSTM to 5 time steps in consideration of the memory limitation and train the network with GT of each sampled frames.

\begin{figure}[t]
	\centering
	\includegraphics[height=6.0cm]{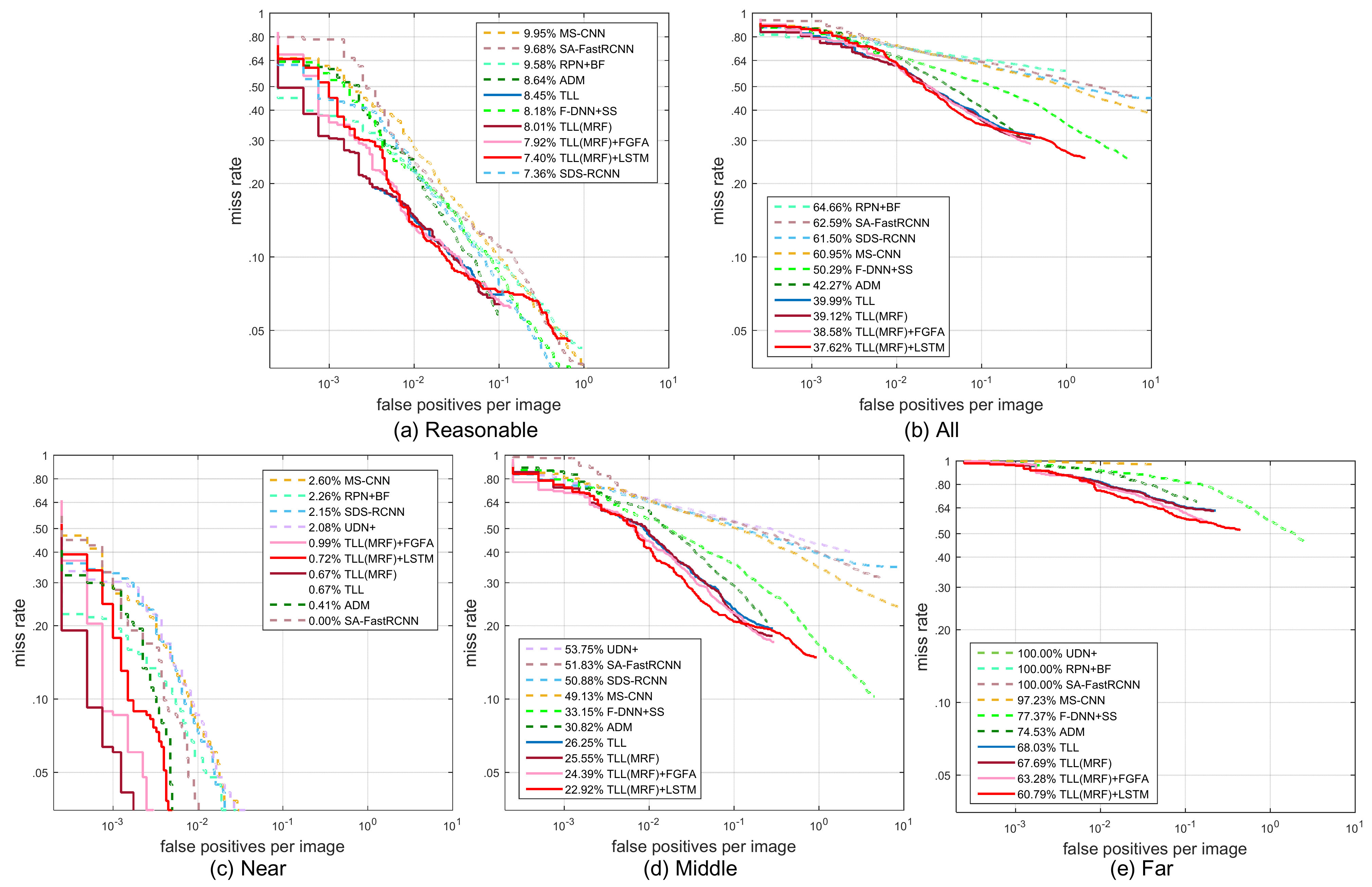}
	\caption{Comparison of our proposed TLL approach with some state-of-the-art methods on the Caltech dataset under Reasonable, All, Near, Middle, and Far evaluation protocols.}
	\label{fig9}
\end{figure}

Fig.~\ref{fig8} illustrates the effect of multi-frame feature aggregation. Columns from left to right are the original image in Caltech test set, the prediction map of topological link edge confidence by single-shot network, and the one by Conv-LSTM based feature aggregation from adjacent frames. It can be seen that for some instances with defocus, blurred boundary and extremely tiny scale, the output feature activations from single-shot network are feeble, or even disappeared. In contrast, Conv-LSTM effectively aggregates the adjacent frame information to the reference frame, resulted in more high-activated features, which benefits the detection of small-scale objects.
Quantitative result of the RNN based TLL is listed in Table~\ref{Cal_res_tab}. Fig.~\ref{fig9} illustrates overall MR--FPPI curves together with best performance benchmarks on the Caltech standard image test set. We also list the result of our TLL combined with FGFA \cite{FGFA}. Compared with FGFA, RNN based aggregation propagates temporal information in a hidden strategy, which allows the network to transfer feature from nearby frames in a more self-driven way, and improves the comprehensive performance more significantly.

\section{Conclusions}
In this work, we design a unified FCN based network to locate the somatic topological line for detecting multi-scale pedestrian instances while introduce a post-processing scheme based on MRF to eliminate ambiguities in occlusion cases. A temporal feature aggregation scheme is integrated to propagate temporal cues across frames and further improves the detection performance. From this work we conclude that: 1) problem itself may reside in the very origin of learning pipeline and it is more appropriate to provide more discriminative and less ambiguous information other than to just feed more information for achieving a better classifier. 2)One should abstract annotations with a more representative methodology. We hope it can inspire more works that focus on intrinsically solving with generic small-scale objects and heavily occlusion scenes.

\bibliographystyle{splncs}
\end{document}